\documentclass[runningheads]{llncs}
\usepackage{graphicx}
\newcommand{\etal}{\textit{et al }.}
\usepackage{amsmath} 
\usepackage{amssymb} 
\usepackage{algorithm}
\usepackage{algorithmic}
\usepackage{here}

\begin{document}
\title{RGB-D Image Inpainting using Generative Adversarial Network \\ with a Late Fusion Approach\thanks{Supported by JST-Mirai Program Grant Number JPMJMI19B2, Japan.}}
%
%
\author{Ryo Fujii\inst{1}\orcidID{0000-0002-9115-8414} \and
Ryo Hachiuma\inst{1}\orcidID{0000-0001-8274-3710} \and \\
Hideo Saito\inst{1}\orcidID{0000-0002-2421-9862}}
\authorrunning{F. Author et al.}
%
\institute{Keio University, Yokohama, Japan \\
\email{\{ryo.fujii0112@,ryo-hachiuma,hs\}@keio.jp}}
\maketitle              
\begin{abstract}
Diminished reality is a technology that aims to remove objects from video images and fills in the missing region with plausible pixels. Most conventional methods utilize the different cameras that capture the same scene from different viewpoints to allow regions to be removed and restored. In this paper, we propose an RGB-D image inpainting method using generative adversarial network, which does not require multiple cameras. Recently, an RGB image inpainting method has achieved outstanding results by employing a generative adversarial network. However, RGB inpainting methods aim to restore only the texture of the missing region and, therefore, does not recover geometric information (i.e, $3$D structure of the scene). We expand conventional image inpainting method to RGB-D image inpainting to jointly restore the texture and geometry of missing regions from a pair of RGB and depth images. Inspired by other tasks that use RGB and depth images (e.g., semantic segmentation and object detection), we propose late fusion approach that exploits the advantage of RGB and depth information each other. The experimental results verify the effectiveness of our proposed method.

\keywords{Image Inpainting \and  Generative Adversarial Network \and Mixed Reality.}
\end{abstract}
\section{Introduction}
Diminished Reality (DR), which allows removing objects from images and filling in the missing regions with plausible textures, plays an important role in many mixed and augmented reality applications. Previous methods for DR can be divided into two groups. The first group takes advantage of multi-view observations to obtain actual background pixels. Pre-observation \cite{5336492}, active self-observation \cite{5543255}, and real-time observation with multiple cameras \cite{1240705} are representative of this group of methods.

They can provide accurate restoration since multi-view-based methods utilize pixels directly observed from other views.  Mori \etal \cite{12aea777c4154aad883678bb1fcb5790}  presented 3D PixMix, which addresses non-planar scenes by using both color and depth information in the inpainting process. However, if the background of the target object is also occluded in other views, a multi-view based method does not work. In the second group, inpainting-based methods can handle this problem, beacause they use pixels in the image to replace pixels that have been removed. Therefore, they do not require multiple cameras and pre-recorded observation.

Image inpainting is the task of synthesizing alternative content in missing regions. It can be used for many applications, such as photo editing, image-based rendering, and computational photography.  Inpainting of RGB image can restore the region's texture, but it cannot restore the geometric structure of missing regions. In this paper, we aim to perform texture and geometry restoration of the missing region.

Traditional image inpainting works make use of low-level features from surrounding images. They work well on background inpainting tasks. However, they are unable to create novel image content not found in the source image. They often fail to restore complex missing regions and objects (e.g., faces) with none-repetitive structures. Moreover, they cannot consider high-level semantics.

Thanks to the rapid development of deep convolutional neural networks  (CNNs) and generative adversarial networks (GANs) \cite{NIPS2014_5423}, convolutional encoder-decoder architectures jointly trained with adversarial networks have been used for inpainting tasks. Iizuka \etal \cite{IizukaSIGGRAPH2017} improved the consistency of image inpainting results by introducing a global and local discriminator. In this paper, we employ this global and local discriminator. 

While GANs-based works show promising visually realistic images, it is quite difficult to make the training of GANs stable. Many methods are proposed on this subject, but stable training of GANs remains unresolved.  Arjovsky \etal \cite{pmlr-v70-arjovsky17a} proposed WGAN  to handle this problem and prevent model collapse. Gulrajani \etal improved WGAN and proposed an alternative way of clipping weight seeking for a more stable training method \cite{NIPS2017_7159}. WGAN-GP has been used for image generation tasks. On this type of task, it is well known that WGAN-GP exceeds the performance of existing GAN losses and works well when combined with $l_1$ reconstruction loss. Yu \etal \cite{Yu2018GenerativeII} proposed to utilize WGAN-GP loss for both outputs of the global and local discriminator. We employ WGAN-GP to make training stable.

One simple solution that restores both the texture and geometry of missing regions is to train two networks independently; one restores textures with an input of RGB image and the other restores geometry with an input from an depth image. We call the no fusion approach. Another possible solution is to train one network which input is RGB-D four-channel image. We call the early fusion approach. Inspired by a recent object recognition method \cite{7258382}, we aim to construct an inpainting network that exploits the complementary relationship between RGB and depth information for RGB-D inpainting. Wang \etal \cite{7258382} showed that the late fusion approach, which combines extracted features from RGB and depth images, improves the classification accuracy of objects in the images. Therefore, we also employ the late fusion approach using RGB and depth information. 

We propose a method to jointly restore texture and geometry information in an image. Our method is based on GAN with the input of a pair of RGB. We employ the late fusion approach to fuse RGB and depth information. The experimental results show that our method successfully restored the missing regions of both RGB and depth images.

Our contributions are as follows:
\begin{itemize}
  \item We propose deep learning architecture that jointly restores the texture and geometry of scenes from RGB and depth images
  \item We employ the late fusion approach to fuse RGB and depth information and show late fusion approach is nicer than early and no fusion approach.
\end{itemize}

The rest of this paper are organized as follows. We first provide preliminaries of our work in Section 2. Section 3 reviews related work on image inpainting. The proposed RGB-D inpainting method is presented in Section 4. Section 5 describes the experiment setting and evaluate results and performance comparisons. This paper is concluded in Section 6. Finally, we discuss the future work in Section 7.

\section{Preliminaries}
\subsection{Generative adversarial networks}
The concept of generative adversarial networks was introduced by Goodfellow \etal \cite{NIPS2014_5423}. Two networks, a discriminator ($D$) and generator ($G$), are jointly trained in the GAN learning process. The generator network learns to map a source of noise to the data space. First, the generator samples input variables from a simple noise distribution such the uniform distribution or spherical Gaussian distribution $P_{z(z)}$, then maps the input variables $z$ to data space $G(z)$. On the other hand, the discriminator network aims to distinguish a generated sample or a true data sample $D(x)$. This relationship can be considered as a minimax two-player game in which $G$ and $D$ compete. The generative and discriminator can be trained jointly by solving the following loss function:
\begin{equation}
\min_{G}\max_{D}\mathbb{E}_{x \sim P_{data(x)}}[\log{D(x)}]+\mathbb{E}_{z \sim P_{z(z)}}[\log{(1-D(G(z)))}]
\end{equation}
\subsection{Wasserstein GANs}
Minimizing the objective  function of GAN is equal to minimizing the Jensen-Shannon divergence between the data and model distributions. GANs are known for their ability to generate high-quality samples, however, training the original version of GANs suffers from many problems (e.g., model collapse and vanishing gradients). To address these problems, Arjovsky \etal \cite{pmlr-v70-arjovsky17a} proposed using the earth-mover (also called Wasserstein-1) distance $W(\mathbb{P}_{g}, \mathbb{P}_{g})$ for comparing the generated and real data distributions as follows:
\begin{equation}
W(\mathbb{P}_{g}, \mathbb{P}_{g}) = \inf_{\gamma \in \prod (\mathbb{P}_{g}, \mathbb{P}_{g})}\mathbb{E}_{(\mathbf{x}, \mathbf{y}) \sim \gamma}[\|\mathbf{x}-\mathbf{y}\|],
\end{equation}
where $\prod (\mathbb{P}_{g}, \mathbb{P}_{g})$ denotes the set of all joint distributions $\gamma(\mathbf{x},\mathbf{y})$ whose marginals are, respectively, $\mathbb{P}_{g}$ and $\mathbb{P}_{g}$.

Its objective function is constructed by applying the Kantorovich-Rubinstein duality:
\begin{equation}
\min_{G}\max_{D\in \mathcal{D}}\mathbb{E}_{\tilde{\mathbf{x}}\sim \mathbb{P}_{g}}[\log{(D(\tilde{\mathbf{x}}))}]-\mathbb{E}_{\textrm{x} \sim \mathbb{P}_{r}}[\log{D(\mathbf{x})}],
\end{equation}
where $\mathcal{D}$ is the set of 1-Lipschitz functions, and $\mathbb{P}_{g}$ is the model distribution, defined by $\tilde{\mathbf{x}} = G(z)$, $z\sim P_z(z)$.

To enforce the Lipschitz constraint in WGAN, Arjovsky \etal added weight clipping to $[-c, c]$. However, recent works suggest that weight clipping reduces the capacity of the discriminator model. Gulrajani \etal proposed an improved version of WGAN adds a gradient penalty term:
\begin{equation}
\lambda\mathbb{E}_{\hat{\mathbf{x}}\sim \mathbb{P}_{\hat{\mathbf{x}}}}[(\|\nabla_{\hat{\mathbf{x}}} D(\hat{\mathbf{x}})\|_2 - 1) ^2],
\end{equation}
where $\hat{\mathbf{x}}$ is sampled from straight lines between pairs of points sampled from the data distribution $\mathbb{P}_{r}$ and the generator distribution $\mathbb{P}_{g}$.

\section{Related Work}
\subsection{Image Inpainting}
In early works, two broad approaches to image inpainting exist. Traditional diffusion-based and patch-based methods belong to the first approach. The diffusion-based method \cite{Ballester:2001:FJI:2318999.2320160} propagates pixel information from around the target missing region in an image. Diffusion-based methods can only fill plain and small or narrow holes. However, the patch-based method, which searches for the closest matching patch and pastes it into the missing region, works well on more complicated images. PatchMatch \cite{Barnes:2009:PAR}, which represents this method, has shown compelling results in practical image editing applications. However, as these methods are heavily based on low-level features (e.g., the sum of squared differences of patch pixel values) and do not consider the global structure, they often cause semantically inconsistent inpainting results. Moreover, they are unable to generate novel objects not found in the existing image.

The second approach is learning-based methods, which train CNNs to predict pixels for missing regions. At the beginning, CNN-based image inpainting approaches can only deal with very small and thin holes and often generate images with artifacts, resulting in blurry and distorted images. To handle these problems, Pathak \etal \cite{pathakCVPR16context} introduced Context Encoder (CE), which is firstly trained with both Mean Square Error (MSE) loss and generative adversarial loss \cite{NIPS2014_5423} as the objective function. This method allows larger mask ($64 \times 64$ mask in a $128 \times 128$ image) to be restored. Iizuka \etal \cite{IizukaSIGGRAPH2017} improves this work by introducing global and local discriminators. The global discriminator judge whether the restored section is consistent with the whole image, and the local discriminator focuses on the inpainted region to distinguish local texture coherency. Furthermore, Iizuka \etal employed dilated convolutions, which allow each layer to increase the area of an input. Yu \etal \cite{Yu2018GenerativeII} proposed an end-to-end image inpainting model consisting of two networks: a coarse network and a refinement network, which ensures the color and texture coherency of generated regions with surroundings. They also introduced a context attention module that allows networks to use information from distant spatial locations and applied a modified version of WGAN-GP loss \cite{NIPS2017_7159} instead of existing GAN loss to ensure global and local consistency and make training stable. We also employ this WGAN-GP in the proposed method.

\subsection{Learning-based RGB-D inpainting}
Only one prior work \cite{Dhamo2019PeekingBO}, which discussed RGB-D image restoration. It focused on predicting depth map and foreground separation mask for hallucinating plausible colors and depths in the occluded area. In this work, they use the independent completion network, a discriminator for RGB and depth images and a pair discriminator for RGB-D image. We employ the idea of pair discriminator in our proposed method.

\begin{figure*}[tb]
 \centering 
 \includegraphics[width = \linewidth]{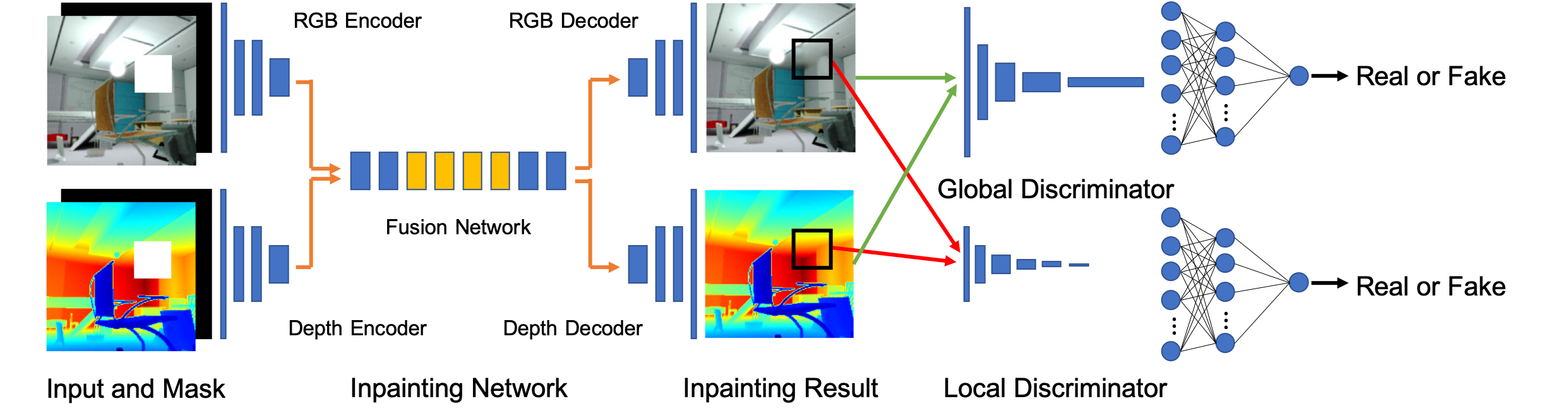}
 \caption{Illustration of the RGB-D image inpainting network. The network takes the input of RGB and depth images with missing regions, and it jointly outputs restored RGB and depth images.}
 \label{fig:network}
\end{figure*}

\section{Approach}
Our proposed method predicts both missing RGB and depth pixels jointly. This approach restores not only the textures but also the geometries of a scene We propose using the GAN-based late fusion architecture to utilize each feature as complementary information. Our network consists of two parts, a completion network and a discriminator network.

\subsection{The Network Architecture}
\subsubsection{ RGB-D Inpainting Network} 
In Figure \ref{fig:network}, our inpainting network is depicted. Our inpainting network is basically based on the architecture proposed upon the Iizuka \etal \cite{IizukaSIGGRAPH2017}. It consists of an RGB encoder-decoder, a depth encoder-decoder, and a fusion network. The inpainting network is applied to images with a rectangle-shaped region with missing pixel. Hole region is implied with white pixels filled in the holes and a binary mask. The hole size and location are randomly selected during training. RGB and depth features, which are extracted from individual encoders, are used as the input of the fusion network. We adopt dilated convolution layers with different dilation rates in the four convolution layers of the fusion part to increase receptive fields.

\subsubsection{ Discriminator Network} 
We employ a discriminator network architecture proposed by Iizuka \etal \cite{IizukaSIGGRAPH2017}. It consists of two networks: a local discriminator and a global discriminator. The global discriminator judges scene consistency, and the local discriminator assesses the quality of the small completed area. Following Dhamo \etal \cite{Dhamo2019PeekingBO} method, to encourage inter-domain consistency between RGB and depth, we set the discriminator's input as RGB-D data, four channels.

\subsection{Loss Function}
We combine content loss and generative adversarial loss for training. Unlike using the standard version of GAN loss, we adopt WGAN-GP, which makes training stable. We use $l_1$ reconstruction loss of RGB and depth images for content loss. WGAN-GP works well when combined with $l_1$ reconstruction as $Wasserstein$-1 distance in the WGAN is based on $l_1$ distance.
\begin{equation}
\mathcal{L}_{content} = \mathcal{L}_{c} + \alpha \mathcal{L}_{d}
\end{equation}
where $\mathcal{L}_{c}$ and $\mathcal{L}_{d}$ denote RGB and depth $l_1$ reconstruction loss, respectively. We set $\alpha$ as $1$.

Given a raw RGB image $\mathbf{x}_c$ and depth image $\mathbf{x}_d$, we choose a random size and location for the binary image mask. We make holes by pixel-wise multiplication of the image and mask ($\mathbf{z}_c=\mathbf{x}_c\odot\mathbf{m}$, $\mathbf{z}_d=\mathbf{x}_d\odot\mathbf{m}$). The RGB encoder and depth encoder takes concatenation of each image and binary mask. Therefore, the input of the RGB encoder and depth encoder are four channels (R, G, B color channels and the binary mask) and two channels (D, depth channels, and the binary mask), respectively. We utilize $[-1, 1]$ normalized image pixels as an input image of the network, and generate an output image with the same resolution $G(\mathbf{z}_c, \mathbf{z}_d, \mathbf{m})$. The details of training procedure are shown in Algorithm \ref{alg1}.

\begin{algorithm}                      
\caption{The training algorithm of our proposed method}         
\label{alg1}                          
\begin{algorithmic}
\WHILE{G has not converged}
    \FOR {$i = 1$ to $5$}
        \STATE Sample batch images $\mathbf{x}_c$, $\mathbf{x}_d$  from training data;
        \STATE Generate random masks $\mathbf{m}$ for $\mathbf{x}_c$, $\mathbf{x}_d$;
        \STATE Construct inputs  $\mathbf{z}_c\leftarrow\mathbf{x}_c\odot\mathbf{m}$, $\mathbf{z}_d\leftarrow\mathbf{z}_d\odot\mathbf{m}$;
       
        \STATE Get predictions 
        \begin{equation*}
            \begin{aligned}
            \tilde{\mathbf{x}}_c&\leftarrow&\mathbf{z}_c+G(\mathbf{z}_c, \mathbf{z}_d, \mathbf{m} )_c\odot(1-\mathbf{m});\\
            \tilde{\mathbf{x}}_d&\leftarrow&\mathbf{z}_d+G(\mathbf{z}_c, \mathbf{z}_d \mathbf{m})_d\odot(1-\mathbf{m});
            \end{aligned}
        \end{equation*}
        \STATE Sample $t\sim U[0,1]$ and $\hat{\mathbf{x}}_{cd}\leftarrow(1-t)\mathbf{x}_{cd}+t\tilde{\mathbf{x}}_{cd}$;
        \STATE Update two critics with $\mathbf{x}_{cd}$,$\tilde{\mathbf{x}}_{cd}$ and $\hat{\mathbf{x}}_{cd}$;
       
    \ENDFOR
    \STATE Sample batch images $\mathbf{x}_c$, $\mathbf{x}_d$ from training data;
    \STATE Generate random masks $\mathbf{m}$ for $\mathbf{x}_c$, $\mathbf{x}_d$;
    \STATE Update inpainting network $G$ with RGB and depth $l1$ reconstruction loss and two adversarial losses;
\ENDWHILE
\end{algorithmic}
\end{algorithm}

\section{Experiments}

\subsection{Dataset}
We evaluate our network with the SceneNet RGB-D dataset \cite{McCormac:etal:ICCV2017}, which consists of five million rendered RGB-D images from over $15,000$ trajectories in synthetic layouts. The poses of objects are randomly arranged and physically simulated with random lighting, camera trajectories, and textures. We train our model using about two million images taken from the SceneNet RGB-D dataset. Following the image inpainting task, we use images of size $256\times256$. We also conduct tests on $10,000$ images from the test data.

\subsection{Implementation details}
  The input of the completion network is an RGB image, depth image, and a binary channel. The outputs are inpainted RGB and depth images. The models are implemented by Pytorch 1.2.0.

\subsection{Baseline methods}
To evaluate our late fusion approach, we also trained two baseline models  using no fusion and early fusion approaches. In the no fusion approach, RGB and depth images are compensated via two individual networks. Therefore, one network uses the RGB image, and the other network uses the depth image as input. While the late fusion network has encoder-decoder architecture for RGB and depth, respectively, the early fusion approach only has encoder-decoder architecture for RGB-D data. For both the no fusion and early fusion approaches, other configurations are the same as that of late fusion.

\subsection{Training Procedure}
The sizes of holes are between $1/8$ and $1/2$ of the size of the image. We train the inpainting network for $900,000$ iterations using a batch size of $32$ images on an NVIDIA Titan GV 100 GPU. The total training time is roughly one week, and the inference is in real-time. We train our model with  Adam optimizer \cite{kingma2014method} with the learning rate of $0.001$.

\begin{figure*}[tb]
\centering
\resizebox{1\textwidth}{!}{\begin{tabular}{ccc|cc|cc}
\begin{minipage}{0.1\hsize}
    \begin{center}
        Input Image
    \end{center}
\end{minipage}
&
\begin{minipage}{0.2\hsize}
    \begin{center}
        \includegraphics[clip, width=\hsize]{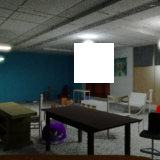} 
    \end{center}
\end{minipage}
&
\begin{minipage}{0.2\hsize}
    \begin{center}
        \includegraphics[clip, width=\hsize]{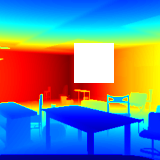} 
    \end{center}
\end{minipage}
&
\begin{minipage}{0.2\hsize}
    \begin{center}
        \includegraphics[clip, width=\hsize]{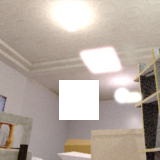} 
    \end{center}
\end{minipage}
&
\begin{minipage}{0.2\hsize}
    \begin{center}
        \includegraphics[clip, width=\hsize]{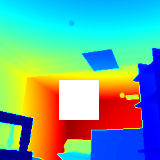} 
    \end{center}
\end{minipage}
&
\begin{minipage}{0.2\hsize}
    \begin{center}
        \includegraphics[clip, width=\hsize]{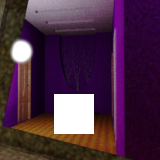} 
    \end{center}
\end{minipage}
&
\begin{minipage}{0.2\hsize}
    \begin{center}
        \includegraphics[clip, width=\hsize]{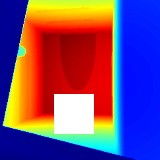} 
    \end{center}
\end{minipage}
\\
\begin{minipage}{0.1\hsize}
    \begin{center}
    Output Image
    \end{center}
\end{minipage}
&
\begin{minipage}{0.2\hsize}
    \begin{center}
        \includegraphics[clip, width=\hsize]{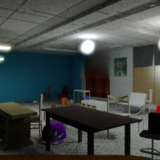} 
    \end{center}
\end{minipage}
&
\begin{minipage}{0.2\hsize}
    \begin{center}
        \includegraphics[clip, width=\hsize]{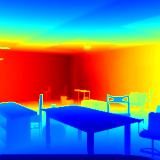} 
    \end{center}
\end{minipage}
&
\begin{minipage}{0.2\hsize}
    \begin{center}
        \includegraphics[clip, width=\hsize]{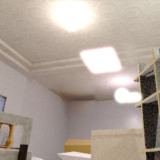} 
    \end{center}
\end{minipage}
&
\begin{minipage}{0.2\hsize}
    \begin{center}
        \includegraphics[clip, width=\hsize]{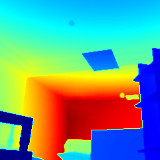} 
    \end{center}
\end{minipage}
&
\begin{minipage}{0.2\hsize}
    \begin{center}
        \includegraphics[clip, width=\hsize]{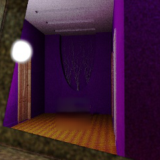} 
    \end{center}
\end{minipage}
&
\begin{minipage}{0.2\hsize}
    \begin{center}
        \includegraphics[clip, width=\hsize]{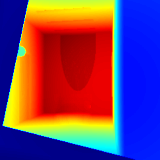} 
    \end{center}
\end{minipage}
\\
\begin{minipage}{0.1\hsize}
    \begin{center}
      Ground Truth
    \end{center}
\end{minipage}
&
\begin{minipage}{0.2\hsize}
    \begin{center}
        \includegraphics[clip, width=\hsize]{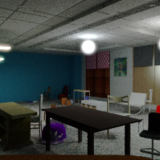} 
    \end{center}
\end{minipage}
&
\begin{minipage}{0.2\hsize}
    \begin{center}
        \includegraphics[clip, width=\hsize]{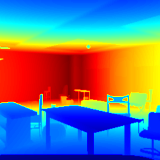} 
    \end{center}
\end{minipage}
&
\begin{minipage}{0.2\hsize}
    \begin{center}
        \includegraphics[clip, width=\hsize]{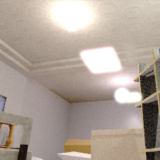} 
    \end{center}
\end{minipage}
&
\begin{minipage}{0.2\hsize}
    \begin{center}
        \includegraphics[clip, width=\hsize]{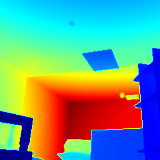} 
    \end{center}
\end{minipage}
&
\begin{minipage}{0.2\hsize}
    \begin{center}
        \includegraphics[clip, width=\hsize]{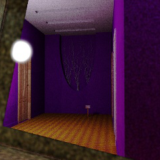} 
    \end{center}
\end{minipage}
&
\begin{minipage}{0.2\hsize}
    \begin{center}
        \includegraphics[clip, width=\hsize]{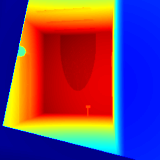} 
    \end{center}
\end{minipage}
\end{tabular}}
\caption{The results of the proposed inpainting network. Missing regions in the images of the first row are colored in white. The pixels with large depth values are colored in red and the pixels with small depth values are colored in blue.}
\label{fig:fig1}
\end{figure*}

\begin{figure*}[tb]
\centering
\resizebox{\textwidth}{!}{
\begin{tabular}{cc|ccc}
\begin{minipage}{0.2\hsize}
    \begin{center}
        Ground Truth
    \end{center}
\end{minipage}
&
\begin{minipage}{0.2\hsize}
    \begin{center}
         Input Image
    \end{center}
\end{minipage}
&
\begin{minipage}{0.2\hsize}
    \begin{center}
        Late fusion (Ours)
    \end{center}
\end{minipage}
&
\begin{minipage}{0.2\hsize}
    \begin{center}
         Early fusion (baseline)
    \end{center}
\end{minipage}
&
\begin{minipage}{0.2\hsize}
    \begin{center} 
        No fusion (baseline)
    \end{center}
\end{minipage}
\\
\begin{minipage}{0.2\hsize}
    \begin{center}
        \includegraphics[clip, width=\hsize]{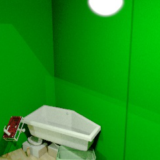} 
    \end{center}
\end{minipage}
&
\begin{minipage}{0.2\hsize}
    \begin{center}
        \includegraphics[clip, width=\hsize]{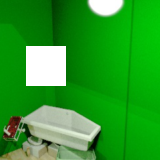} 
    \end{center}
\end{minipage}
&
\begin{minipage}{0.2\hsize}
    \begin{center}
        \includegraphics[clip, width=\hsize]{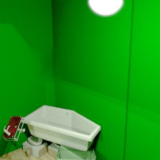} 
    \end{center}
\end{minipage}
&
\begin{minipage}{0.2\hsize}
    \begin{center}
        \includegraphics[clip, width=\hsize]{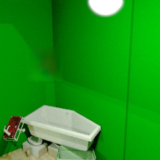} 
    \end{center}
\end{minipage}
&
\begin{minipage}{0.2\hsize}
    \begin{center}
        \includegraphics[clip, width=\hsize]{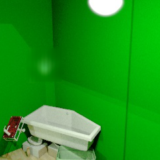} 
    \end{center}
\end{minipage}
\\
\begin{minipage}{0.2\hsize}
    \begin{center}
        \includegraphics[clip, width=\hsize]{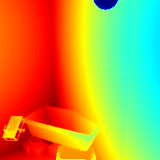} 
    \end{center}
\end{minipage}
&
\begin{minipage}{0.2\hsize}
    \begin{center}
        \includegraphics[clip, width=\hsize]{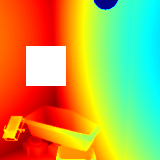} 
    \end{center}
\end{minipage}
&
\begin{minipage}{0.2\hsize}
    \begin{center}
        \includegraphics[clip, width=\hsize]{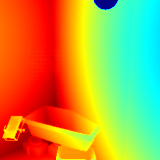} 
    \end{center}
\end{minipage}
&
\begin{minipage}{0.2\hsize}
    \begin{center}
        \includegraphics[clip, width=\hsize]{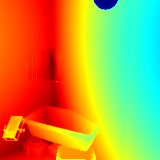} 
    \end{center}
\end{minipage}
&
\begin{minipage}{0.2\hsize}
    \begin{center}
        \includegraphics[clip, width=\hsize]{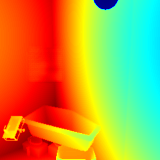} 
    \end{center}
\end{minipage}
\\ 
\begin{minipage}{0.2\hsize}
    \begin{center}
        \includegraphics[clip, width=\hsize]{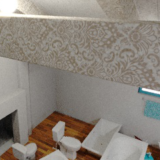} 
    \end{center}
\end{minipage}
&
\begin{minipage}{0.2\hsize}
    \begin{center}
        \includegraphics[clip, width=\hsize]{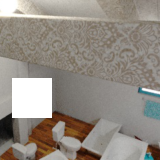} 
    \end{center}
\end{minipage}
&
\begin{minipage}{0.2\hsize}
    \begin{center}
        \includegraphics[clip, width=\hsize]{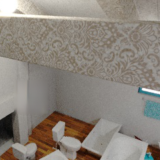} 
    \end{center}
\end{minipage}
&
\begin{minipage}{0.2\hsize}
    \begin{center}
        \includegraphics[clip, width=\hsize]{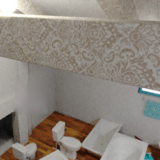} 
    \end{center}
\end{minipage}
&
\begin{minipage}{0.2\hsize}
    \begin{center}
        \includegraphics[clip, width=\hsize]{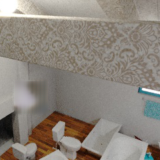} 
    \end{center}
\end{minipage}
\\
\begin{minipage}{0.2\hsize}
    \begin{center}
        \includegraphics[clip, width=\hsize]{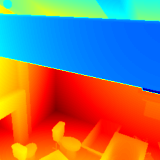} 
    \end{center}
\end{minipage}
&
\begin{minipage}{0.2\hsize}
    \begin{center}
        \includegraphics[clip, width=\hsize]{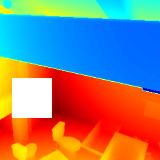} 
    \end{center}
\end{minipage}
&
\begin{minipage}{0.2\hsize}
    \begin{center}
        \includegraphics[clip, width=\hsize]{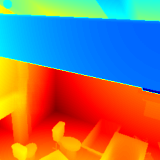} 
    \end{center}
\end{minipage}
&
\begin{minipage}{0.2\hsize}
    \begin{center}
        \includegraphics[clip, width=\hsize]{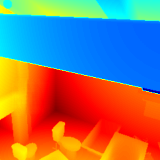} 
    \end{center}
\end{minipage}
&
\begin{minipage}{0.2\hsize}
    \begin{center}
        \includegraphics[clip, width=\hsize]{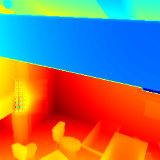} 
    \end{center}
\end{minipage}
\end{tabular}
}

\caption{Qualitative comparisons of our proposed method with two baselines.}
\label{fig:fig2}
\end{figure*}

\begin{figure}[tb]
\begin{center}
\resizebox{\textwidth}{!}{\begin{tabular}{ccc}
\begin{minipage}{0.15\hsize}
    \begin{center}
       Input Image
    \end{center}
\end{minipage}
&
\begin{minipage}{0.15\hsize}
    \begin{center}
        Output Image
    \end{center}
\end{minipage}
&
\begin{minipage}{0.15\hsize}
    \begin{center}
        Ground Truth
    \end{center}
\end{minipage}
\\
\begin{minipage}{0.3\hsize}
    \begin{center}
        \includegraphics[clip, width=\hsize]{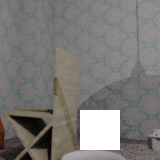} 
    \end{center}
\end{minipage}
&
\begin{minipage}{0.3\hsize}
    \begin{center}
        \includegraphics[clip, width=\hsize]{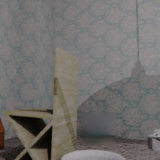}
    \end{center}
\end{minipage}
&
\begin{minipage}{0.3\hsize}
    \begin{center}
        \includegraphics[clip, width=\hsize]{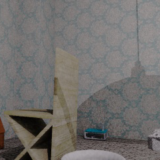} 
    \end{center}
\end{minipage}
\\
\begin{minipage}{0.3\hsize}
    \begin{center}
        \includegraphics[clip, width=\hsize]{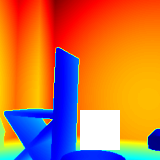}
    \end{center}
\end{minipage}
&
\begin{minipage}{0.3\hsize}
    \begin{center}
        \includegraphics[clip, width=\hsize]{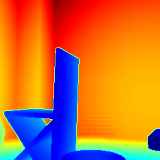} 
    \end{center}
\end{minipage}
&
\begin{minipage}{0.3\hsize}
    \begin{center}
        \includegraphics[clip, width=\hsize]{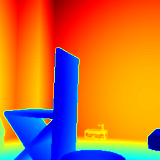} 
    \end{center}
\end{minipage}
\end{tabular}}
\caption{The result of the proposed method with the input of the object mask. }
\label{fig:fig3}
\end{center}

\end{figure}

\subsection{Performance Evaluation}
\subsubsection{Qualitative Comparison}
In Figure \ref{fig:fig1}, we show the inpainting result of three images from the dataset. We confirmed that our network restores the missing region of both RGB and depth images. This shows that the late fusion approach formed the edge of each restored region clearly.

We compare the qualitative performance of our late fusion approach with early and no fusion approaches. As shown in Figure \ref{fig:fig2}, compared to baseline methods, late fusion successfully merges the depth clue into RGB results to enhance sharp edges. However, early fusion sometimes provides a discontinuous result. The no fusion approach shows obvious visual artifacts, including blurred or distorted images in the masked region in both RGB and the depth of the masked region.

Moreover, we use the RGB-D image masked to the object as the input of our proposed network. As depicted in Figure \ref{fig:fig3}, our method generates a plausible depth inpainted image, while the RGB outcome includes some blurs. This result shows the possibility of applying our method to diminished reality (DR) applications.
\subsubsection{Quantitative Comparison}

\subsubsection{ RGB inpainting result}
Table \ref{tab:rgbq} shows the quantitative results of the RGB restoration of our proposed method. We present a comparison with baseline methods to verify the effectiness of our proposed method. Since many possible solutions different from the original image content exist, image inpainting task does not have perfect metrics. Nevertheless, we evaluate our proposed method by measuring  $l1$ loss, signal-to-noise ratio (PSNR) and the structural similarity (SSIM) following the RGB inapainting task \cite{IizukaSIGGRAPH2017}. Our proposed method slightly outperforms two baseline methods.
\renewcommand{\arraystretch}{1.5}
\begin{table}[tb]
\centering
\caption{RGB quantitative result.}
\begin{tabular}{lcccc}
  \hline
 Method & $l_1$$\downarrow$ & PSNR$\uparrow$ & SSIM$\uparrow$ \\
   \hline
Early Fusion & $4.27 \times 10^{-3}$ &  $\mathbf{3.94 \times 10^{-3}}$ & $0.985$  &  \\ 
No Fusion &  $6.33 \times 10^{-3}$ &  $2.75 \times 10^{-3}$ & $0.980$  & \\ 
Late Fusion (Ours) & $\mathbf{3.38 \times 10^{-3}}$    &  $2.61 \times 10^{-3}$  &   $\mathbf{0.987}$  &  \\ 
   \hline
\end{tabular}

\label{tab:rgbq}
\end{table}

\subsubsection{Depth inpainting result}
Table \ref{tab:depthq} illustrates the quantitative evaluation of the depth inpainting result. We compute the absolute relative error (Abs rel), squared relative error (Sq rel), root mean square error (RMSE), and logged root mean square error (RMSE log) following the monocular depth estimation task \cite{Eigen2014DepthMP}. From the table, our proposed method significantly improves the accuracy of the depth inpainting.

\begin{table}[tb]
\centering
\caption{Depth quantitative result.}
\begin{tabular}{lcccccc}
  \hline
 Method & $l_1$ $\downarrow$ & Abs Rel$\downarrow$ & Sq Rel$\downarrow$ & RMSE$\downarrow$ & RMSE log$\downarrow$ &\\ 
  \hline

Early Fusion & $11.0$ & $4.29 \times 10^{-3}$ & $10.1$ & $77.2$ & $2.02 \times 10^{-2}$ & \\ 
No Fusion & $14.3$ & $4.06 \times 10^{-3}$ & $5.45$ & $93.4$ & $2.28 \times 10^{-2}$ & \\ 
Late Fusion (Ours) & $\mathbf{6.83}$ & $\mathbf{2.06 \times 10^{-3}}$ & $\mathbf{2.21}$ & $\mathbf{52.2}$ & $\mathbf{1.29\times10^{-2}}$ & \\
 
   \hline
\end{tabular}
\label{tab:depthq}
\end{table} 

\section{Conclusion}
In this work, we proposed a GAN-based RGB-D encoder-decoder inpainting network and evaluated late fusion on a synthetic dataset. Our network jointly restored the missing region of RGB and depth images. We discussed the fusion method enhancing each inpainted result complementary. We showed that the late fusion approach outperforms no fusion and early fusion approaches.

\section{Future Work}
First, we will compare our proposed method with \cite{Dhamo2019PeekingBO}, which also uses NYU depth v2 (real) for evaluation, in addition to SceneNet \cite{McCormac:etal:ICCV2017} (synthetic). 
Second, we will explore the fusion method. In this method, we show the advantage of the late fusion approach, of which fusion way is very simple. Fusion schemes for RGB-D image inpainting have not been proposed, but they have been explored for other tasks. Zeng \etal \cite{Qiu_2019_CVPR} focused on surface normal estimation from RGB-D data using a hierarchical network with adaptive feature re-writing, which are proposed to fuse color and depth features at multiple scales. In their work, Zeng \etal stated that single-scale fusion (our method) is inefficient for fusing RGB and depth when RGB and depth contain different type of noise.

Lastly, we will also study the validation of the discriminator. In this paper, we focus on the architecture of the generator and do not pay much attention to the discriminator model. Discriminators for depth images have not been studied thoroughly. Following the monocular depth estimation task, we will propose a discriminator that judges inpainted depth at multiple scales.

%
%
%
\bibliographystyle{splncs04}
\bibliography{mybib}

\end{document}